\documentclass{article}

\usepackage{cite}
\usepackage{amsmath,amssymb,amsfonts}
\usepackage{algorithmic}
\usepackage{graphicx}
\usepackage{textcomp}
\usepackage{amsmath,amssymb,amsfonts}
\usepackage{xcolor}
\usepackage{multirow}
\usepackage{rotating}
\usepackage{subcaption}
\usepackage{hyperref}
\usepackage{booktabs}
\usepackage{fancyhdr}
\usepackage{float}
\usepackage{threeparttable}
\usepackage{makecell} 

\usepackage[numbers]{natbib}

\usepackage{arxiv}

\usepackage[utf8]{inputenc} 
\usepackage[T1]{fontenc}    
\usepackage{hyperref}       
\usepackage{url}            
\usepackage{booktabs}       
\usepackage{amsfonts}       
\usepackage{nicefrac}       
\usepackage{microtype}      
\usepackage{lipsum}		
\usepackage{graphicx}
\usepackage{amsmath,amssymb,amsfonts}
\usepackage{natbib}
\usepackage{doi}
\usepackage{multicol}

\usepackage{threeparttable}
\usepackage{makecell}

\title{Real-Time Currency Detection and Voice Feedback for
 Visually Impaired Individuals}

\author{
Saraf Anzum Shreya\textsuperscript{1}\thanks{Email: sasshreya2001@gmail.com}, 
MD. Abu Ismail Siddique\textsuperscript{1}\thanks{Email: saif101303@gmail.com}, 
Sharaf Tasnim\textsuperscript{1}\thanks{Email: sharaftasnim786@gmail.com} \\
\textsuperscript{1}Dept. of Electronics and Telecommunication Engineering, \\Rajshahi University of Engineering and Technology, Rajshahi, Bangladesh
}

\renewcommand{\headeright}{}
\renewcommand{\undertitle}{}
\renewcommand{\shorttitle}{Real-Time Currency Detection and Voice Feedback for
 Visually Impaired Individuals}

\hypersetup{
pdftitle={A template for the arxiv style},
pdfsubject={q-bio.NC, q-bio.QM},
pdfauthor={David S.~Hippocampus, Elias D.~Striatum},
pdfkeywords={Currency Detection,  Computer Vision, Deep Learning, Machine Learning and YOLO},
}

\begin{document}
\maketitle

\begin{abstract}
	Technologies like smartphones have become an essential in our daily lives. It has made accessible to everyone including visually impaired individuals. With the use of smartphone cameras, image capturing and processing have become more convenient. With the use of smartphones and machine learning, the life of visually impaired can be made a little easier. Daily tasks such as handling money without relying on someone can be troublesome for them. For that purpose this paper presents a real-time currency detection system designed to assist visually impaired individuals. The proposed model is trained on a dataset containing 30 classes of notes and coins, representing 3 types of currency: US dollar (USD), Euro (EUR), and Bangladeshi taka (BDT). Our approach uses a YOLOv8 nano model with a custom detection head featuring deep convolutional layers and Squeeze-and-Excitation blocks to enhance feature extraction and detection accuracy. Our model has achieved a higher accuracy of 97.73\%, recall of 95.23$\%$, f1-score of 95.85$\%$ and a mean Average Precision at IoU=0.5 (mAP50(B)) of 97.21$\%$. Using the voice feedback after the detection would help the visually impaired to identify the currency. This paper aims to create a practical and efficient currency detection system to empower visually impaired individuals independent in handling money.
\end{abstract}
\maketitle

\section{Introduction}
\label{sec:introduction}
Throughout the history, money has always been an essential component of human societies. Currency is the main accepted exchange tool across different cultures and economies. People rely on currency for daily activities whether they use paper money, coins or digital payment methods. Currency helps people to buy necessary items such as food, clothing as well as pay for transportation, healthcare services and other essential needs. Managing money effectively is important for gaining personal independence, building better financial knowledge and staying actively involved in social and community life.

The global financial landscape shows that physical cash remains as a fundamental method of transactions despite advancements in digital payment technology. Cash remains the dominant transaction method in developing regions with limited digital infrastructure. Physical currency remains vital for small businesses and rural communities because they lack access to electronic payment systems and banking facilities. Physical cash remains a main component of financial systems around the world. In advanced digital economies, cash still remains the preferred payment method for specific transactions because it delivers dependable accessibility with personal financial control. 

For visually impaired individuals, the ability to independently identify and handle currency is essential for maintaining financial autonomy and participating fully in society. While most modern banknotes incorporate security features such as watermarks, holograms, and microprinting, these are primarily designed to prevent counterfeiting rather than assist those with visual impairments. A large number of the population is in fact visually impaired. In 2020, 43·3 million were estimated to be blind. Estimated 295 million people to have moderate and severe vision impairment\cite{GBD_2019_Blindness_and_Vision_Impairment_Collaborators2021-ci}. According to Dhaka Tribune (2018), approximately 750k people in Bangladesh are blind\cite{noauthor_undated-bj}. Using the power of technology disabled people can be self-dependent. In the daily lives of visually impaired persons, they are dependent on others for most tasks, for one handling money and transactions. 

Tactile markings on banknotes such as raised dots or embossed patterns, are designed to assist visually impaired individuals in identifying currency denominations. However, their effectiveness depends heavily on awareness and familiarity with these features. Only about 10$\%$ of the visually impaired know to read and write tactile system \cite{article}. The rest of the 90$\%$ may not recognize or understand the purpose of these tactile markings. For those people, it is nearly impossible to distinguish between different denominations. Even those who know tactile, face difficulties identifying currency. Feedback from Australian currency users showed that people with less sensitivity to textile markings have a harder time identifying currencies. Also paper notes lose the tactile marking's sensitivity and effectiveness due to constant wear and tear\cite{RePEc:bca:bcarev:v:2010:y:2010:i:winter09-10:p:32-39}. 

A camera-based detection or identification of currency can be a more efficient alternative. It addresses many of the limitations of tactile systems such as wear and tear providing a more versatile and user-friendly solution. Despite the challenges like technological dependences and recognition errors, ongoing improvement in machine learning and universal access to technology will most likely to produce more effective and popularized systems in the future.

A study\cite{Martiniello2022-qa} on people with visual impairment using smartphones revealed that more than 90$\%$ among the respondents used smartphones for activities such as calls, sending and receiving messages, browsing the web and reading emails. Around 70–80$\%$ used smartphones for calendar functions, listening to music, social media and networking. By integrating a machine learning model into a mobile application, visually impaired individuals could benefit from real-time accurate currency recognition. Smartphone-based detection ensures consistency regardless of a note’s condition. The use of digital image processing has made easier to detect and identify currency under various conditions such as poor lighting, water and tear and folded notes.

A smartphone-based approach can identify multiple currencies without requiring the users to memorize different tactile patterns. It is possible to update and improve recognition models through software updates. This makes the system adaptable to new banknote designs. The system also uses real-time audio feedback to ensure faster identification without the need to see or feel for the tactile features. It not only enhances independence for visually impaired individuals but also makes currency handling faster, more efficient, and accessible in various real-world scenarios.

\begin{figure*}[t]
    \centering
    \includegraphics[width=1\textwidth]{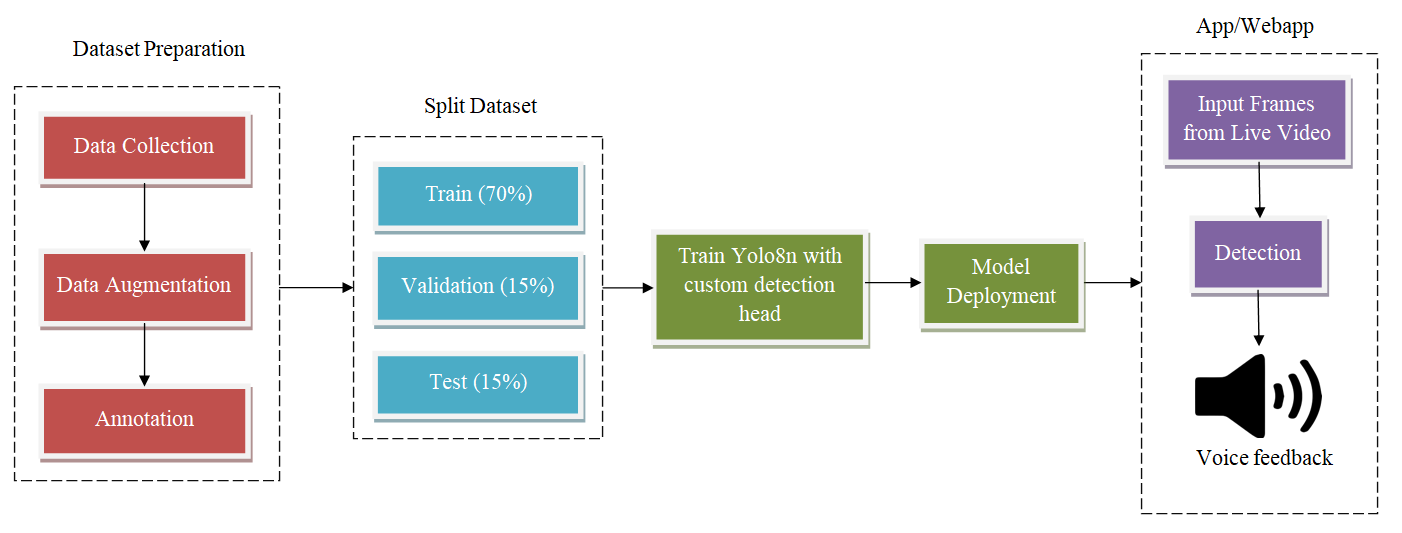}
    \caption{Overflow of the Process}
    \label{fig:research-roadmap}
\end{figure*}

\section{Literature Review}

Several studies have explored different methods of currency identification for the visually impaired. Almost every currency has tactile markings as a way for the visually impaired to detect the notes. But their effectiveness is limited by how familiar users are with them and the wear on the currency. It makes the challenge even greater since many visually impaired individuals don’t have easy access to Braille literacy. To eliminates these challenges there have been efforts to use machine learning to help detect currency in real-time. While these systems offer better accuracy and convenience, they still face some challenAges like improving the model’s performance and ensuring that the technology is accessible to everyone.

The process of the production of the banknotes are complex. It goes through the offset printing, screen printing, and intaglio printing. It creates detailed and secure features like tactile patterns for the visually impaired. These features are important for security but it can be time-consuming especially manual engraving. In 1988 Polymer banknotes were introduced which offer increased durability and resistance to counterfeiting, even though challenges persist in creating effective recognition marks for visually impaired users\cite{tactile}. 

In a paper by Abu Doush and AL-Btoush \cite{AbuDoush2016}, a currency recognition system using a smartphone camera was developed that compares color Scale-Invariant Feature Transform (SIFT) and grayscale SIFT algorithms. The system was tested on a dataset of Jordanian banknotes and coins captured under different conditions. The color SIFT approach outperforms the grayscale SIFT in both recognition accuracy and processing time. The color SIFT method achieved higher correct recognition rates especially for paper currencies and processing times suitable for mobile deployment.

Han and Kim \cite{HanKim2019} proposes a joint banknote recognition and counterfeit detection system using convolutional neural networks (CNN) coupled with explainable artificial intelligence techniques. The proposed system simultaneously performs on the US Dollar and Euro banknotes denomination recognition and counterfeit detection, significantly improving processing speed. The method incorporates a novel pixel-wise Grad-CAM (pGrad-CAM) visualization to explain the model’s decision process utilizing visible, infrared transmission and infrared reflection images. The results show 100\% accuracy in recognition and counterfeit detection with around 8ms inference time which outperforms sequential approaches.

The study by Park et al.\cite{9217508} suggested a deep learning-based approach to help visually impaired individuals identify banknotes and coins using caremas of smartphones. A three-stage framework was suggested with Faster R-CNN for initial detection. Then false positives were removed using geometric constraints like width-to-height ratios and detection scores. The final stage classifies the objects into coins, bills, or mixed categories using ResNet-based Faster R-CNN. The model was tested on the Dongguk Korean Banknote (DKB) and Jordanian Dinar (JOD) databases. Despite having high accuracy, the system struggles detecting severely folded or occluded banknotes

The paper by Singh et al.\cite{9868010}, presents "IPCRF", an end-to-end framework for Indian paper currency recognition tailored for blind and visually impaired people. The proposed IPCRNet model is a lightweight CNN model combined with MobileNet as the front-end with a novel Contextual Block (CB) in the back-end. It uses dense connections and multi-dilation depth-wise separable convolutions. This design enhances feature extraction while maintaining a low parameter count (3.6M) which is suitable for resource-constrained devices. The authors also introduced "IPCD", a large-scale dataset of 50,263 images with diverse scenarios like folded and partial views to better simulate real-world usage. The model outperforms state-of-the-art networks achieving 96.75$\%$ accuracy on IPCD. They have also developed an app called "Roshni" for real-time currency recognition.

The paper\cite{math11061392} introduces a deep learning-based framework that helps visually impaired individuals in recognizing banknotes from multiple nationalities using smartphone cameras. The MBDM employs mosaic data augmentation to enhance training data diversity which helps improve detection accuracy. They have trained the model on a dataset of currencies such as Korean won (KRW), US dollars (USD), Euros (EUR) and Jordanian dinar (JOD). The model achieved an accuracy of 0.8396, recall of 0.9334 and F1 score of 0.8840. Their model outperforms state-of-the-art methods like YOLO-v3 and Faster R-CNN. Its main contributions include a 69-layer CNN optimized for detecting small objects like coins, mosaic augmentation for improved generalization and a publicly available multinational banknote database (DMB v1). While performance is slightly lower for smaller coins due to their size and reflective surfaces, the MBDM represents a significant advancement in assistive technology, offering a reliable solution for multinational banknote recognition.

The paper\cite{8639124} presents a deep learning-based approach for currency detection and recognition using the Single Shot MultiBox Detector (SSD) framework combined with Convolutional Neural Networks (CNN). The authors have used a small dataset of New Zealand currency denominations (5NZD, 10NZD, 20NZD) containing 300 images. The they augmented it to 7,500 images. The model achieved an impressive accuracy of 96.6$\%$ by extracting features from both the front and back sides of the banknotes. The study shows the effectiveness of SSD and CNN in accurately detecting and classifying currency in spite of different angles and lighting.

On the other hand, the paper \cite{9868010} suggests an implementation method of the IPCRNet model in a mobile application named "Roshni-Currency Recognizer". The lightweight model is optimized for smartphones, is compressed into TensorFlow Lite (TFLite) format for efficient offline use. The app uses the smartphone's camera to capture currency images to processes them with IPCRNet and provides voice-based feedback to announce the denomination. It includes features like auto-start, voice guidance and customizable settings which sets them apart from other suggested systems. They focus on a mobile app rather than a web application ensuring offline functionality.

The papers \cite{9217508}, \cite{HanKim2019}, \cite{AbuDoush2016} and \cite{math11061392} do not explicitly mention the implementation of their currency recognition systems.  These papers present robust methods for currency recognition but they do not provide details on implementing these systems. The deployment aspects may be left for future work or practical applications.

\section{Materials \& Methods}
\subsection{Overflow of Proposed Method}
The figure [\ref{fig:research-roadmap}] shows the roadmap of our proposed method. The dataset were collected from various sources to build a comprehensive dataset. These images were gathered from real-world scenarios to ensure diversity and relevance. Data augmentation was applied after collecting the raw images to increase the dataset's size and variability. This step is crucial for improving the model's ability to generalize and perform well in different conditions.

After the augmentation the dataset underwent annotations. Each image was labeled with bounding boxes where the whole area of the note was annotated. Additionally for more accuracy the numbers and texts were annotated. Roboflow\cite{roboflow} was used for this purpose as they streamline the annotation process and ensure high-quality labeled data. The annotated dataset was then split into training and validation sets for model training and evaluation.

With the prepared dataset the next step was to train the models. In this case, the proposed model was acquired with a custom detection head with yolo backbone and neck. Before training with this model, the data was preprocessed to highlight the features and edges. The proposed model was then compared with the standard YOLOv8 nano model. The standard YOLOv8 nano model is trained using the annotated images optimizing its parameters to minimize detection errors. Our proposed model incorporates additional neural network layers in the detection head and data preprocessing is used to improve feature extraction and detection accuracy. The proposed model was evaluated on the test sets to assess their performance.

Finally, the trained models are deployed for real-world implementation. The proposed model is directly integrated into a web-view based application. The deployment involves using Flask for the backend and a combination of HTML, JavaScript and WebSocket communication for the frontend. The application captures live video frames from the user's camera then processes them in the server using the proposed model and provides real-time detection results along with voice feedback.

\subsection{Data Acquisition}
A dataset of 3 types of currency, such as the US dollar (USD), Euro (EUR) and Bangladeshi taka (BDT) was collected. For the US dollars, images were collected from a publicly available dataset, "usd bill classification dataset"\cite{Techie2023}  as well as stack images. Only the desired images were taken after carefully choosing the desired images. There were used 7 denominations for USD (1 dollar, 2 dollar, 5 dollar, 10 dollar, 20 dollar, 50 dollar, 100 dollar). Only notes were considered for US dollar and coins were avoided.

\begin{figure}
    \centering
    \includegraphics[width=.95\linewidth]{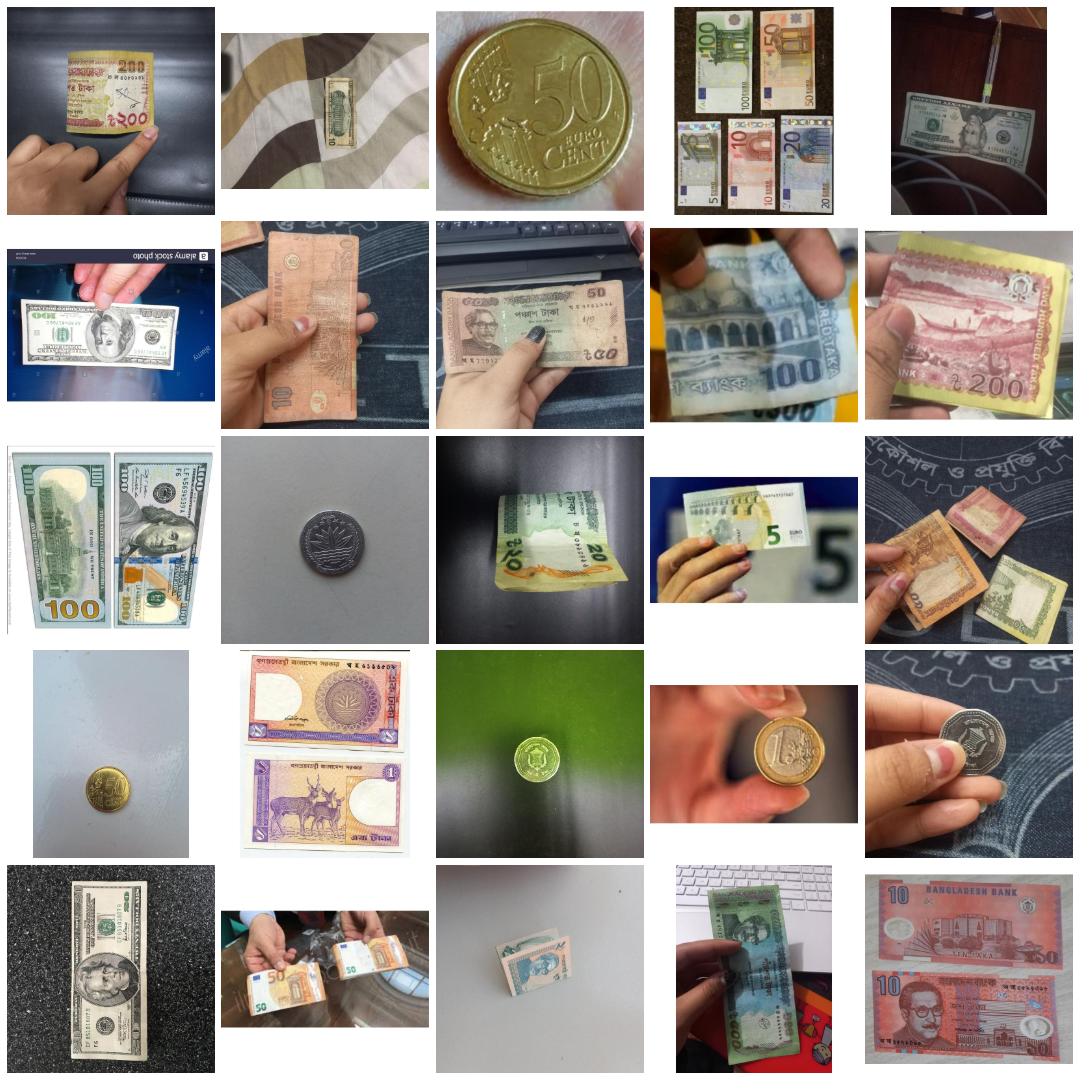}
    \caption{Sample images from the dataset with different backgrounds and lightings.}
    \label{fig:dataset sample}
\end{figure}

Two public datasets were considered for the Euro, "Euro currency"\cite{Miencus2023} and "euro-counter-dataset"\cite{euro-counter_dataset}. And for the BDT, most of the images were taken by an android camera under different lighting and background, some images were taken from the dataset called "thesis dataset"\cite{thesis-osbek_dataset}. Coins were also considered for EUR and BDT.

The BDT dataset has rare notes and coins including the ones that had been discontinued, such as the golden 1 taka coin and 1 taka note. The dataset includes all the updated currencies as well as the old ones. It also has new notes and worn out old notes for generalization.

To make the dataset more diverse, the images were taken with the notes both folded and unfolded, and sometimes only part of the note is visible. Figure [\ref{fig:dataset sample}] shows a few samples from the dataset with different angles, perspectives, lightings and backgrounds.

\subsection{Data Preparation}
The images were augmented and annotated before training a model on the dataset. All the images were made square-shaped by adding white padding and were resized to 640x640 pixels. Several augmentation techniques were applied to enhance the dataset's generalization ability.


The images were at first rotated by 30° and 60°. Then the original and rotated images were further rotated by 90°, 180°, 270°, and 360° to capture a variety of orientations. The reason for applying these augmentation techniques is to simulate different perspectives and angles since the images in the original dataset were captured from only a single angle. This helps to generalize the dataset and help expose the model to a wider range of potential real-world scenarios where the currency notes and coins may appear from various orientations. The dataset expanded to a total of 25,012 images across all the classes after the augmentation. This variety in the dataset allows the model to learn better representations of the objects which improves its performance and generalization capabilities.

\begin{figure}
    \centering
    \includegraphics[width=.95\linewidth]{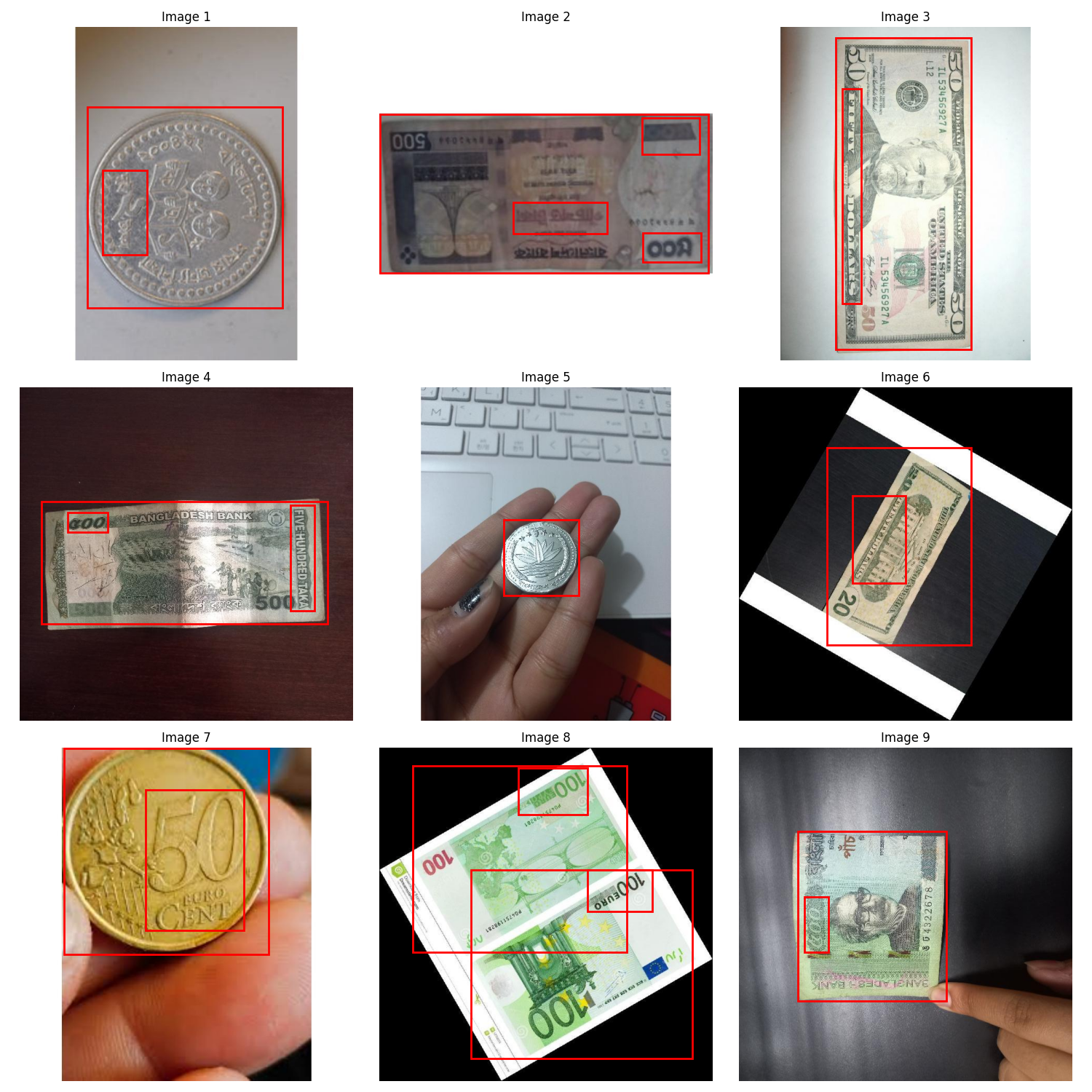}
    \caption{Sample images from the dataset with annotations.}
    \label{fig:annotated_images}
\end{figure}

All augmented images were uploaded to Roboflow \cite{roboflow} for manual annotation. Annotations were carefully added to each image ensuring that both the currency and any distinguishing features, such as numbers and text were marked. Common English numbers and texts were excluded, except for those indicating the currency type (e.g., Taka, Dollar, or Euro), to ensure the uniqueness and avoid redundancy of each currency. This approach ensures that the annotations remained relevant and specific to the unique characteristics of each currency note or coin. The annotations were then saved as .txt files in the YOLO format for easy integration with the training pipeline.

Figure [\ref{fig:annotated_images}] shows a few example images on how the annotations were done.

Before the annotations were done, the dataset was divided into three subsets: 70\% for training, 15\% for validation and 15\% for test. The dataset contains images of various currencies with different denominations of Taka, Dollar, and Euro. The annotations cover both the currency itself and the distinguishing numbers and texts that help identify each note or coin. These annotations were carefully created by hand to ensure accuracy particularly for unique currency identifiers, while avoiding common English text that is typically found across all currency notes. The dataset is made public on Kaggle \cite{saraf_anzum_shreya_2025}.

\begin{table}[ht]
\centering
\renewcommand{\arraystretch}{1.2} 
\setlength{\tabcolsep}{6pt} 
\begin{tabular}{|c|c|c|c|}
\hline
\textbf{Class Name} & \textbf{Train Images} & \textbf{Test Images} & \textbf{Validation Images} \\
\hline
1000taka      & 1888  & 408  & 340  \\\hline
100dollar     & 1912  & 392  & 408  \\\hline
100euro       & 1252  & 268  & 388  \\\hline
100taka       & 1636  & 132  & 408  \\\hline
10dollar      & 1404  & 200  & 232  \\\hline
10euro        & 1068  & 328  & 272  \\\hline
10eurocent    & 944   & 176  & 176  \\\hline
10taka        & 1468  & 300  & 192  \\\hline
1dollar       & 1360  & 276  & 216  \\\hline
1euro         & 412   & 108  & 100  \\\hline
1eurocent     & 460   & 108  & 80   \\\hline
1taka         & 1040  & 152  & 420  \\\hline
200taka       & 936   & 156  & 288  \\\hline
20dollar      & 1148  & 432  & 300  \\\hline
20euro        & 1164  & 208  & 200  \\\hline
20eurocent    & 904   & 220  & 248  \\\hline
20taka        & 2048  & 540  & 388  \\\hline
2dollar       & 3176  & 636  & 560  \\\hline
2euro         & 308   & 48   & 48   \\\hline
2eurocent     & 360   & 112  & 144  \\\hline
2taka         & 2508  & 348  & 304  \\\hline
500taka       & 2736  & 720  & 492  \\\hline
50dollar      & 1628  & 440  & 464  \\\hline
50euro        & 1336  & 220  & 172  \\\hline
50eurocent    & 1084  & 192  & 216  \\\hline
50taka        & 1680  & 368  & 444  \\\hline
5dollar       & 1536  & 332  & 360  \\\hline
5euro         & 1360  & 292  & 328  \\\hline
5eurocent     & 468   & 72   & 108  \\\hline
5taka         & 3488  & 884  & 744  \\\hline
\end{tabular}
\caption{Train, Test, and Validation Instance Counts per Class}
\label{tab:dataset}
\end{table}\vspace{-0.4mm}

\subsection{Data Preprocessing}

The preprocessing pipeline was designed to improve the model's ability for detection and classification by enhance=ing the quality of the input images. The preprocessing steps include Contrast Limited Adaptive Histogram Equalization (CLAHE), sharpening and Gaussian blurring. They were applied to the images before feeding them into the model for training and testing. These steps are important for improving the model's ability to detect fine details and edges in the currency images, especially under varying lighting conditions.

\subsubsection{\textbf{Gaussian Blurring}}
The Gaussian blur is applied to reduce noise in the image and creates a smoother result. It is done by averaging the pixel values in a neighborhood around each pixel using a Gaussian function. The Gaussian blur can be mathematically expressed as\cite{6044249}:

\begin{equation} 
I_{\text{blurred}}(x, y) = \sum_{i,j} I(x+i, y+j) \cdot G(i,j)
\end{equation}

where \( G(i,j) \) is the Gaussian kernel and \( I(x+i, y+j) \) are the pixel values of the image. The result is a blurred image that helps reduce high-frequency noise.

In the preprocessing pipeline of our model, a Gaussian blur with a kernel size of 5×5 is applied to the original image. I was done smoothing the images before further processing.

\subsubsection{\textbf{CLAHE (Contrast Limited Adaptive Histogram Equalization)}}
CLAHE is a technique that is used to enhance the contrast of images. It is done by redistributing the intensity values in localized regions of the image. As it helps to highlight the textures and patterns on images, it is particularly useful for currency detection. The CLAHE algorithm works by dividing the image into small regions called tiles and applying histogram equalization to each region. Then the contrast is limited to avoid amplifying the noise in the image. Mathematically CLAHE can be described as follows\cite{zuiderveld1994contrast}:

\begin{equation}
I_{\text{CLAHE}}(x, y) = \text{CLAHE}(I(x, y))
\end{equation}

where \( I(x, y) \) is the original image and \( I_{\text{CLAHE}}(x, y) \) is the contrast-enhanced image.

The input BGR image is converted to the LAB color space before applying CLAHE in order to separate the image into lightness (L) and color (A and B) components. The lightness channel is isolated for contrast enhancement using CLAHE. CLAHE was initialized with a clip limit of 5.0 to control contrast amplification and an 8×8 tile grid for localized histogram equalization. CLAHE enhances the local contrast of the lightness channel which improves detail visibility while preventing noise over-amplification. The enhanced lightness channel is then merged back with the original color channels. Then the image is converted back to the BGR color space. The final output is an image with improved local contrast and preserved color balance whcih makes the details more prominent.

\subsubsection{\textbf{Image Sharpening}}
Image sharpening is an important step to enhance the fine details and edges. A sharpening filter is applied to the image to highlight the edges and reduce the blurriness after the Gaussian Blur and CLAHE was applied. The sharpening can be mathematically described as follows:

\begin{equation} 
I_{\text{sharpened}}(x, y) = \text{filter}(I_{\text{enhanced}}(x, y), \text{sharpening kernel})
\end{equation}

where \( I_{\text{sharpened}}(x, y) \) is the sharpened image.

A kernel is used to enhance the sharpness of the image with the central value being 5 and the surrounding values being -1. The function applies the filter using OpenCV’s filter2D method and returns the sharpened image, which will help to detect fine currency details.

\subsubsection{\textbf{Preprocessing Pipeline}}
The complete preprocessing pipeline combines the Gaussian blur, CLAHE and sharpening steps in sequence. Initially, the image was denoised using a Gaussian blur. Then CLAHE was applied to improve the contrast of the image to enhance the features. Finally, sharpening was applied to enhance the edges.

The pipeline can be expressed as:

\begin{equation} 
I_{\text{processed}} = \text{sharpen}(\text{enhance}(\text{CLAHE}(\text{blur}(I))))
\end{equation}

where \( I_{\text{processed}} \) is the final processed image.

\begin{figure}
    \centering
    \includegraphics[width=.75\linewidth]{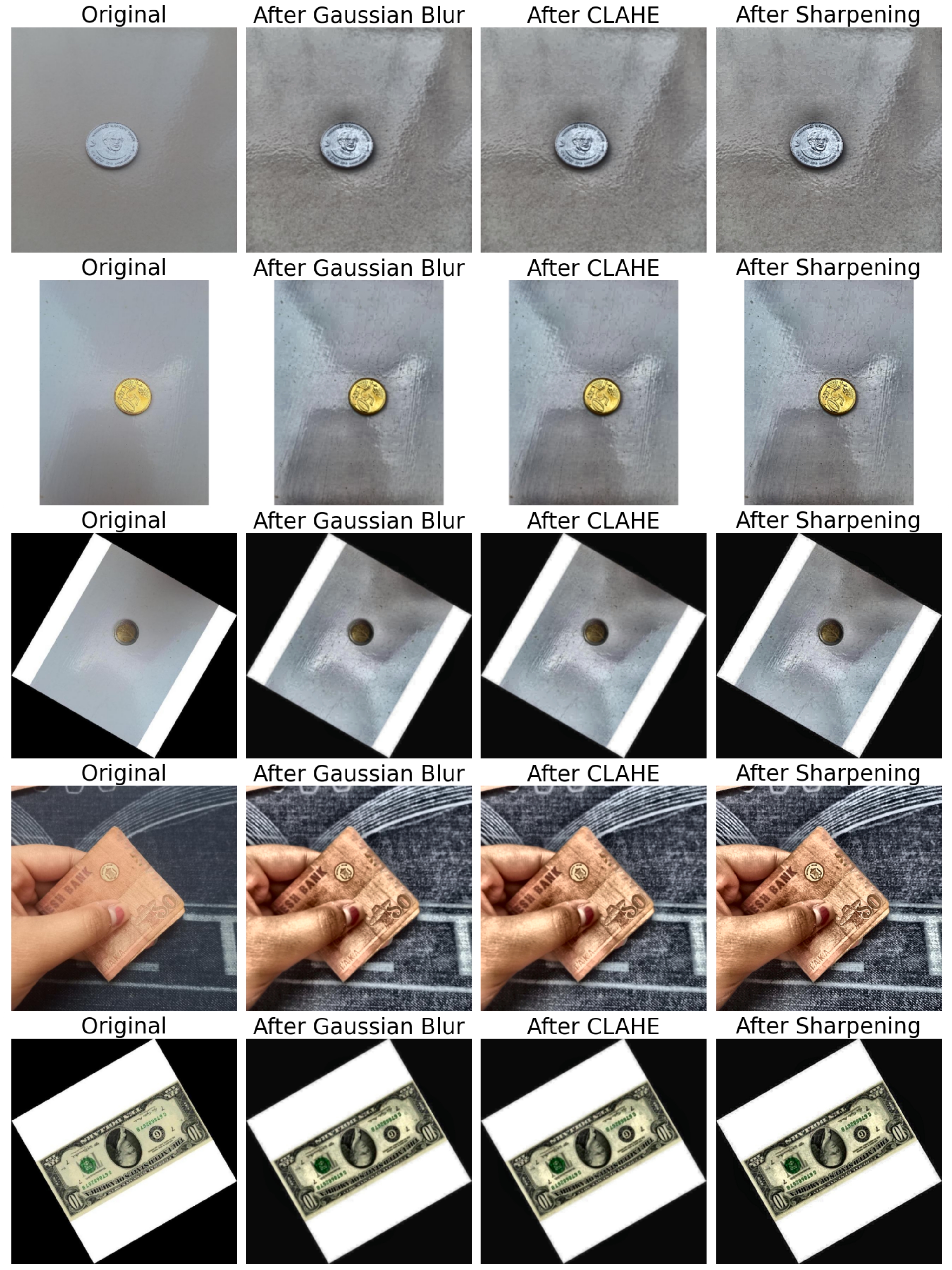}
    \caption{Sample images of the preprocessing pipeline showing (from left to right) the original image, Gaussian blurred image, CLAHE-enhanced image and sharpened image.}
    \label{fig:filter_example}
\end{figure}

\subsection{Model Architecture}
Our proposed model is built upon the YOLOv8n architecture \cite{YOLOv8n}, a compact variant of the YOLOv8 framework developed by Ultralytics. YOLO (You Only Look Once) is a family of single-stage object detection models that predicts bounding boxes and class probabilities directly from an input image in a single forward pass. It offers a compelling trade-off between speed and accuracy compared to the two-stage detectors such as Faster R-CNN \cite{Ren2015FasterRCNN}. The YOLOv8n variant is specifically optimized for efficiency. It is ideal for real-time applications or resource-constrained environments because of its reduced parameter count and computational complexity while preserving robust detection capabilities.

The architecture of YOLOv8n has three primary components: the \textit{backbone}, the \textit{neck} and the \textit{head}. The backbone extracts multi-scale feature maps from the input image by capturing hierarchical representations of visual information which ranges from low-level edges to high-level semantic content. These feature maps are then processed by the neck to aggregate features across different scales using a Feature Pyramid Network (FPN) \cite{Lin2017FPN} or Path Aggregation Network (PAN) \cite{Liu2018PAN} which enhances the model's ability to detect objects of varying sizes. Finally, the head generates predictions including bounding box coordinates, objectness scores and class probabilities based on the aggregated features. The input images are standardized to a resolution of 640$\times$640 pixels to align with the model's default configuration.

\subsubsection{Custom Modifications to the Model}
In this paper we introduce a custom detection head tailored to the specific demands of currency detection. The YOLOv8n backbone and neck components are kept in their original form. The custom detection head replaces the default YOLOv8n head to optimize performance for this application. We incorporated a Squeeze-and-Excitation (SE) block \cite{Hu2018SE} to enhance feature representation. This was done to identify and classify currency notes and coins (e.g., USD, EUR, BDT) from our augmented dataset, necessary for precise localization and classification across a diverse set of object appearances and scales.

\paragraph{Squeeze-and-Excitation (SE) Block}
The SE block\cite{Hu2018SE} is a lightweight attention mechanism designed to recalibrate channel-wise feature responses. It improves the representational power of convolutional neural networks. It operates in two stages: \textit{squeeze} and \textit{excitation}.

The structure of the SE block is detailed in Table [\ref{tab:se_block}].
\begin{table}[ht]
    \centering
    \renewcommand{\arraystretch}{1.2} 
    \setlength{\tabcolsep}{6pt} 
    \begin{threeparttable}
    \begin{tabular}{|l|c|c|}
        \hline
        \textbf{Layer Type} & \textbf{Input Size} & \textbf{Output Size} \\\hline
        AdaptiveAvgPool2d   & \([B, C, H, W]\)    & \([B, C, 1, 1]\)    \\ \hline
        Linear              & \([B, C]\)          & \([B, C/16]\)       \\ \hline
        ReLU                & \([B, C/16]\)       & \([B, C/16]\)       \\ \hline
        Linear              & \([B, C/16]\)       & \([B, C]\)          \\ \hline
        Sigmoid             & \([B, C]\)          & \([B, C, 1, 1]\)    \\ \hline
        Scaling             & \([B, C, H, W]\)    & \([B, C, H, W]\)    \\ \hline
    \end{tabular}
    \begin{tablenotes}
        \item \textit{Note}: \(B\) is the batch size, \(H\) and \(W\) are the height and width of the input feature map, and \(C\) is the number of channels (e.g., 32). The Sigmoid output is broadcast and multiplied element-wise with the original \([B, C, H, W]\) input to the SEBlock.
    \end{tablenotes}
    \caption{Structure of the Squeeze-and-Excitation (SE) Block}
    \label{tab:se_block}
    \end{threeparttable}
\end{table}

In the squeeze phase, global spatial information is aggregated using adaptive average pooling, producing channel descriptors of size $[B, C, 1, 1]$. The excitation phase employs two linear layers with a reduction ratio of 16 (mapping $32 \rightarrow 2 \rightarrow 32$ channels): first reducing dimensionality to $[B, C/16]$, applying ReLU activation, then restoring to $[B, C]$. A sigmoid activation generates channel weights in $[0,1]$, which rescale the original features through element-wise multiplication. This adaptive recalibration enhances the network's focus on discriminative features critical for currency detection, particularly the fine-scale patterns and security features characteristic of banknotes.

\paragraph{Custom Detection Head}
The custom detection head processes feature maps from three distinct scales—P3 (80$\times$80), P4 (40$\times$40), and P5 (20$\times$20)—produced by the YOLOv8n neck which corresponds to shallow, mid-level, and deep features, respectively. These scales enable the detection of objects ranging from small coins to larger notes. The head is designed to predict bounding boxes, objectness scores, and class probabilities for 30 currency classes, with three anchors per spatial location to account for varying aspect ratios and sizes.

For each scale, the detection head applies a sequence of operations: a 3$\times$3 convolutional layer reduces the input channels (64, 128, or 512, depending on the scale) to a uniform 32 channels, followed by batch normalization and a ReLU activation to stabilize training and introduce non-linearity. The SE block then recalibrates the resulting feature maps, enhancing channel-wise feature importance. Finally, a 1$\times$1 convolutional layer generates the output tensor with 105 channels per spatial location and calculated as $3 \times (30 + 5)$, where 3 is the number of anchors, 30 is the number of classes and 5 represents the bounding box parameters (x, y, width, height) and objectness score.

\begin{figure}
    \centering
    \includegraphics[width=0.6\linewidth]{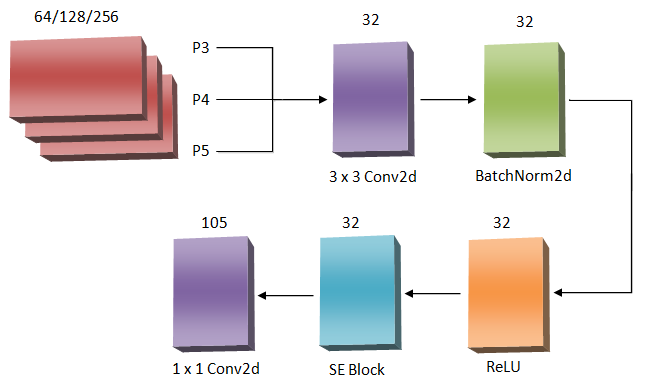}
    \caption{Structure of the Custom Detection Head.}
    \label{fig:custom_head}
\end{figure}

The structure of the custom detection head per scale is presented in figure [\ref{fig:custom_head}].

The overall modified YOLOv8n architecture is summarized in Table [\ref{tab:overall_arch}].
\begin{table*}[ht]
    \centering
    \renewcommand{\arraystretch}{1.2}
    \begin{threeparttable}
    \begin{tabular}{|l|l|c|c|}
        \hline
        \makecell{\textbf{Component}} & 
        \makecell{\textbf{Description}} & 
        \makecell{\textbf{Input Size}} & 
        \makecell{\textbf{Output Size} \\\textbf{(per scale)}} \\
        \hline
        \makecell{Backbone}           & \makecell{Default \\YOLOv8n}      & \makecell{640$\times$640$\times$3} & \makecell{P3: 80$\times$80$\times$64,\\ P4: 40$\times$40$\times$128,\\ P5: 20$\times$20$\times$256} \\\hline
        \makecell{Neck}               & \makecell{FPN/PAN \\(retained)}   & \makecell{Multi-scale}            & \makecell{P3: 80$\times$80$\times$64,\\ P4: 40$\times$40$\times$128,\\ P5: 20$\times$20$\times$256} \\\hline
        \makecell{Detection\\Head}    & \makecell{CustomDetect\\Head}     & \makecell{P3/P4/P5}               & \makecell{\texttt{[B,105,80,80]}, \\ \texttt{[B,105,40,40]}, \\ \texttt{[B,105,20,20]}} \\ 
        \hline
    \end{tabular}
    \end{threeparttable}
    \caption{Overall Modified YOLOv8n Architecture}
    \label{tab:overall_arch}
\end{table*}

\begin{figure*}[t]
    \centering
    \includegraphics[width=0.95\textwidth]{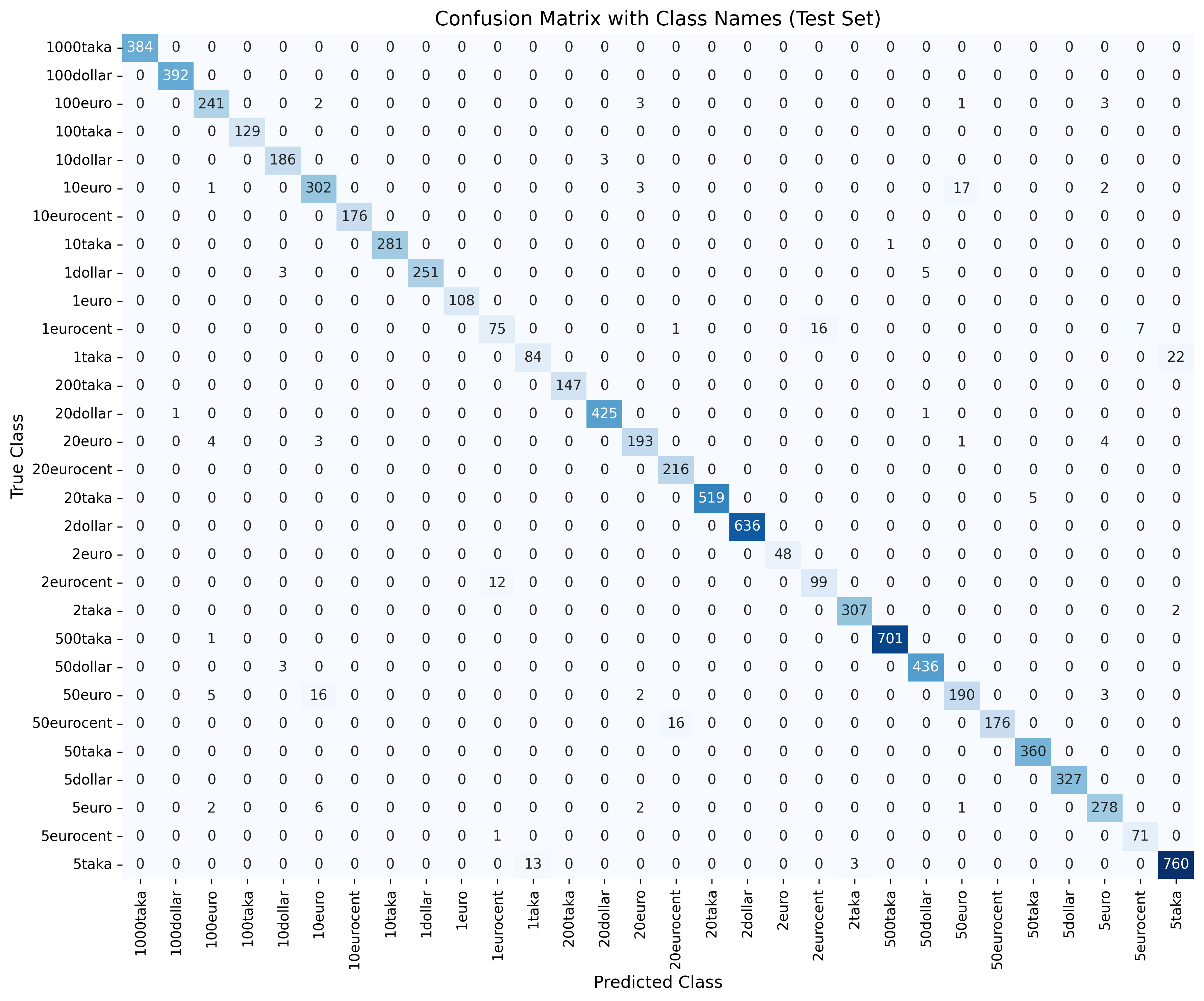}
    \caption{Confusion matrix for the proposed model across 30 currency classes.}
    \label{fig:confusion_matrix}
\end{figure*}

\subsubsection{Training Methodology}
The model is trained using the Stochastic Gradient Descent (SGD) optimizer with an initial learning rate of 0.01 and a batch size of 64. Training spans 50 epochs on the USD-EUR-BDT dataset, with images resized to 640$\times$640 pixels. Model performance is evaluated on a test set using metrics such as precision, recall, and mean Average Precision (mAP) at IoU thresholds of 0.5 (mAP@50) and 0.5:0.95 (mAP@50:95), ensuring a comprehensive assessment of detection accuracy and robustness.

The integration of the SE block within the custom detection head enhances the model's ability to focus on critical features, such as the fine details of currency notes. The multi-scale design ensures effective detection across varying object sizes. This architecture keeps a balance between computational efficiency and detection performance, making it well-suited for our currency detection task.

\subsection{Evaluation Metrics}

\begin{figure*}
    \centering
    \includegraphics[width=1\textwidth]{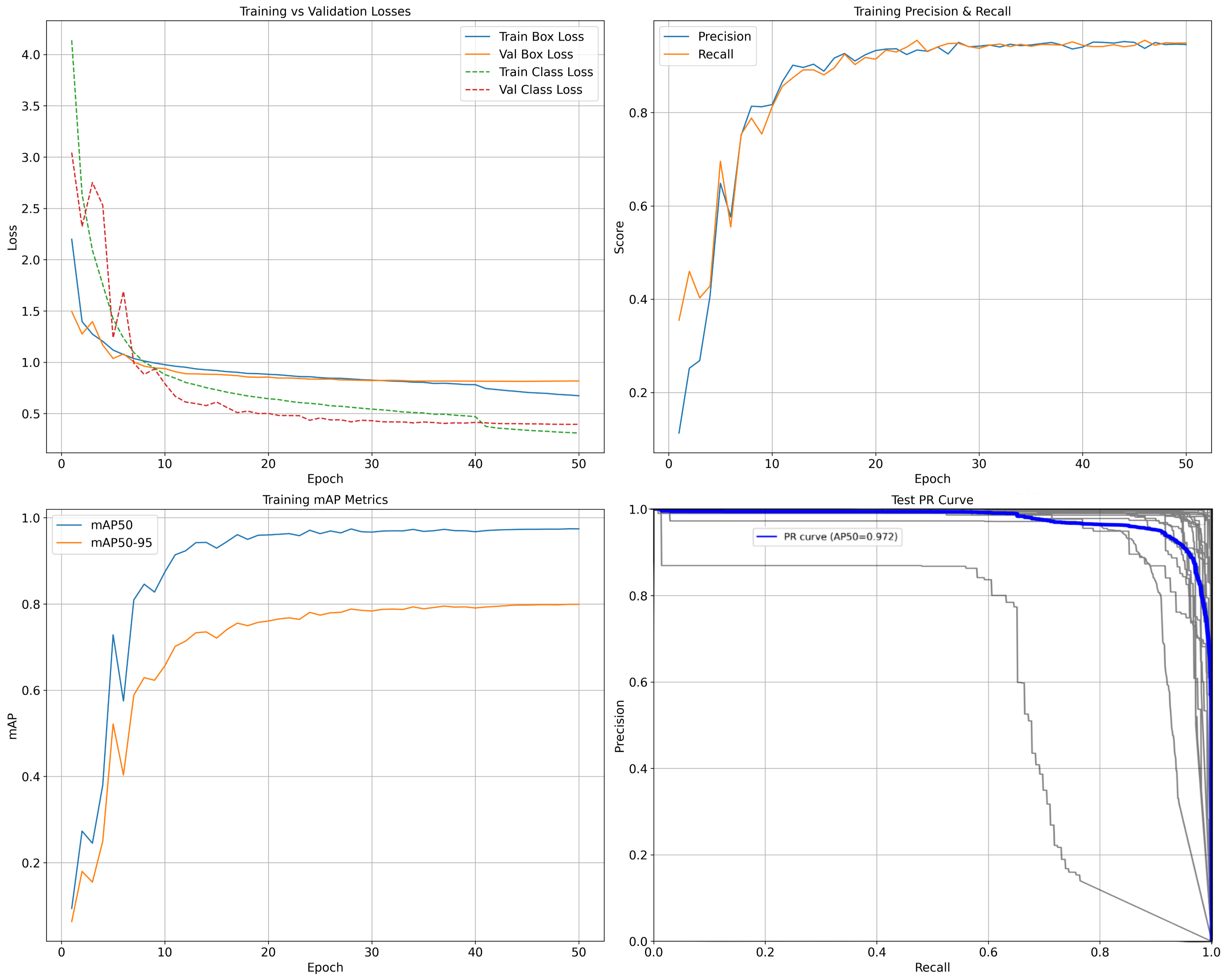}
    \caption{Evaluation metrics curves for the proposed model over 50 epochs.}
    \label{fig:matrics_curve}
\end{figure*}

The performance of the model was evaluated using several key metrics, including accuracy, precision, recall, F1-score, and mean Average Precision (mAP). These metrics are commonly used in object detection tasks to assess the model's ability to identify and localize objects in an image correctly.

\subsubsection{Accuracy}
Accuracy is a metric that measures the overall correctness of the model’s predictions. It is defined as the ratio of correct predictions (true positives and true negatives) to the total number of predictions:

\begin{equation}
\text{Accuracy} = \frac{\text{TP} + \text{TN}}{\text{TP} + \text{TN} + \text{FP} + \text{FN}}
\end{equation}

\subsubsection{Precision}
Precision measures the accuracy of the model's positive predictions. It is defined as the ratio of true positives (TP) to the sum of true positives and false positives (FP):

\begin{equation}
\text{Precision} = \frac{\text{TP}}{\text{TP} + \text{FP}}
\end{equation}

\subsubsection{Recall}
Recall measures the model's ability to identify all relevant instances. It is defined as the ratio of true positives (TP) to the sum of true positives and false negatives (FN):

\begin{equation}
\text{Recall} = \frac{\text{TP}}{\text{TP} + \text{FN}}
\end{equation}

\subsubsection{F1-Score}
The F1-score provides a balanced measure of precision and recall, making it particularly useful in scenarios where false positives and false negatives are equally important. It is defined as the harmonic mean of precision and recall:

\begin{equation}
\text{F1-Score} = 2 \times \frac{\text{Precision} \times \text{Recall}}{\text{Precision} + \text{Recall}}
\end{equation}

\subsubsection{Mean Average Precision (mAP)}
The mAP metric evaluates the model's overall performance across all currency classes by taking into account both the accuracy of the predictions and the model's ability to localize the objects correctly. The mAP is calculated as the average of the precision values at different recall levels:

\begin{equation}
\text{mAP} = \frac{1}{N} \sum_{i=1}^{N} \text{AP}_i
\end{equation}

where \( N \) is the number of classes (30 in this study), and \( \text{AP}_i \) is the average precision for class \( i \).

\subsection{Results \& Analysis}

\subsubsection{Confusion Matrix Analysis}
The model's performance was evaluated using a confusion matrix. It provides a detailed breakdown of predictions across the 30 currency classes (e.g., "1000taka," "100dollar," "50euro," etc.). Figure [\ref{fig:confusion_matrix}] illustrates the confusion matrix, which reveals high accuracy for most classes, with strong diagonal values (e.g., 384 for "1000taka," 392 for "100dollar"), reflecting minimal confusion. However, some classes like "5euro" (190 true positives, 16 confused with "10euro"), show minor misclassifications, likely due to visual similarities in design or denomination proximity. This analysis confirms the model's robustness while highlighting areas for refinement in distinguishing closely related classes.

\subsubsection{Loss Analysis}
The training and validation losses for both box and class predictions are shown in the top-left graph of Figure [\ref{fig:matrics_curve}]. The Train Box Loss (green) and Val Box Loss (blue) start high and decrease sharply. It stabilizes around 0.5-1.0 after 10-20 epochs which indicates effective learning of bounding box predictions with good generalization. The Train Class Loss (red) and Val Class Loss (orange) follow a similar trend, dropping from approximately 3.0 to 0.5-1.0 which demonstrates accurate object classification. 

\subsubsection{Precision and Recall curves}
The top-right graph of Figure [\ref{fig:matrics_curve}] presents the precision and recall curves over 50 epochs. Precision (blue) increases rapidly and stabilizes near 0.95, indicating fewer false positives as training progresses. Recall (orange) also rises sharply and stabilizes at a similar level which suggests that the model effectively detects most objects. This alignment confirms the model learns without significant overfitting.

\subsubsection{mAP Metrics Analysis}
The bottom-left graph of Figure [\ref{fig:matrics_curve}] shows the Training mAP Metrics. The mAP50 (blue) and mAP50-95 (orange) increase steadily, reaching approximately 0.9 and 0.8 respectively by epoch 50. This reflects strong mean Average Precision across IoU thresholds (0.5 and 0.5-0.95). It indicates robust object detection and localization performance.

\subsubsection{PR AUC Curve}
The bottom-right graph of Figure [\ref{fig:matrics_curve}] displays the Test PR Curve (blue) with AP50=0.972, showing a high area under the curve. This indicates an excellent balance of precision and recall on the test set with the curve remaining near the top-left corner, confirming high accuracy and effectiveness in real-world scenarios.


\subsubsection{F-1 score and accuracy Evaluation}
The proposed model achieved a precision of 96.47\%, indicating high accuracy in identifying correct currency denominations with minimal false positives, and a recall of 95.23\%, demonstrating effectiveness in detecting most true positives across the 30 classes, as validated by Figure [\ref{fig:matrics_curve}]. The F1-score, balancing precision and recall, was calculated as:

\begin{equation}
\text{F1-Score} = 2 \times \frac{0.9647 \times 0.9523}{0.9647 + 0.9523} = 0.9585 \text{ (95.85\%)}
\end{equation}

This 95.85\% F1-score reflects a robust balance, crucial for real-time currency detection aiding visually impaired users, where both missed detections and false alarms affect usability.

The accuracy of the model was calculated from the confusion matrix using the following formula:

\begin{equation}
\text{Accuracy} = \frac{\text{True Positives} + \text{True Negatives}}{\text{Total Predictions}} 
\end{equation}
\begin{equation}
\text{Accuracy} = \frac{\text{Trace of the Confusion Matrix}}{\text{Sum of All Elements in the Confusion Matrix}}
\end{equation}

For the proposed model, the calculated accuracy is 97.73\%, while the YOLOv8 nano model achieved 97.64\%. This shows a slight improvement in the proposed model's ability to correctly classify currency denominations, providing enhanced performance for real-time currency detection.

The inference speed was 3.1 ms per image (0.2 ms preprocess, 2.2 ms inference, 0.7 ms postprocess), achieved with YOLOv8n, making it ideal for real-time deployment on resource-constrained devices like smartphones.

\subsubsection{Comparison with Standard YOLOv8n}
The proposed model was compared with standard YOLOv8n trained on the same USD-EUR-BDT dataset, as shown in Table [\ref{tab:comparison}]. Our model outperforms YOLOv8n in accuracy (97.73\% vs. 97.64\%), recall (95.23\% vs. 94.36\%), F1-score (95.85\% vs. 95.57\%), and mAP@0.5 (97.21\% vs. 96.97\%), driven by a custom detection head with Squeeze-and-Excitation (SE) blocks enhancing currency-specific feature focus. It also achieves a faster total time of 3.1 ms vs. 3.2 ms, due to a 0.1 ms preprocessing reduction from an optimized pipeline.

YOLOv8n slightly excels in precision (96.80\% vs. 96.47\%) and mAP@0.5:0.95 (80.65\% vs. 80.38\%), reducing false positives and improving strict localization. However, for assisting visually impaired users, higher recall is critical to minimize missed detections, making our model’s trade-off preferable. The confusion matrix (Figure [\ref{fig:confusion_matrix}]) supports this, showing high true positives across classes, with minor errors not significantly impacting usability when paired with voice feedback.

\begin{table*}[ht]
    \centering
    \renewcommand{\arraystretch}{1.2}
    \begin{threeparttable}
    \begin{tabular}{|l|c|c|}
        \hline
        \makecell{\textbf{Metric}} & \makecell{\textbf{YOLOv8n}} & \makecell{\textbf{Proposed Model}} \\
        \hline
        Accuracy              & 97.64\%           & \textbf{97.73\%}                    \\\hline
        Precision (B)             & \textbf{96.80\%}           & 96.47\%                    \\\hline
        Recall (B)                & 94.36\%                    & \textbf{95.23\%}           \\\hline
        F1-Score                  & 95.57\%                    & \textbf{95.85\%}           \\\hline
        mAP@0.5 (B)               & 96.97\%                    & \textbf{97.21\%}           \\\hline
        mAP@0.5:0.95 (B)          & \textbf{80.65\%}           & 80.38\%                    \\\hline
        Preprocess (ms)           & 0.3                        & \textbf{0.2}               \\\hline
        Inference (ms)            & 2.2                        & 2.2                        \\\hline
        Postprocess (ms)          & 0.7                        & 0.7                        \\\hline
        Total Time (ms)           & 3.2                        & \textbf{3.1}               \\
        \hline
    \end{tabular}
    \end{threeparttable}
    \caption{Performance Comparison: YOLOv8n vs. Proposed Model}
    \label{tab:comparison}
\end{table*}

\subsubsection{Comparison with Recent Studies}

To evaluate the efficacy of the proposed model, we compare it with three recent studies on currency recognition: Park and Park (2023)\cite{math11061392}, Singh et al. (2022)\cite{9868010}, and Park et al. (2020)\cite{9217508}. These studies represent diverse approaches including multinational banknote detection, lightweight CNN frameworks, and multi-stage detection with Faster R-CNN. Table [\ref{tab:recent_comparison}] summarizes the key aspects of each study alongside our proposed YOLOv8n-based model.

\begin{table*}[ht]
    \centering
    \renewcommand{\arraystretch}{1.2}
    \begin{threeparttable}
    \begin{tabular}{|p{2.5cm}|p{3cm}|p{3cm}|p{3cm}|p{3cm}|}
        \hline
        \makecell{Study} & \makecell{Park and Park (2023)\cite{math11061392}} & \makecell{Singh et al. (2022)\cite{9868010}} & \makecell{Park et al. (2020)\cite{9217508}} & \makecell{Proposed Model} \\
        \hline
        \makecell{Method} & 
        Multinational banknote detection model (MBDM) based on 69-layer CNN with mosaic data augmentation & 
        Lightweight CNN (IPCRNet) using MobileNet front-end and contextual block with multi-dilation depthwise separable convolutions & 
        Three-stage Faster R-CNN detection with geometric constraints and ResNet-18 verification & 
        YOLOv8 nano with custom detection head and Squeeze-and-Excitation blocks \\
        \hline
        \makecell{Dataset} & 
        Korean won, US dollar, Euro, Jordanian dinar; 21,020 images from multiple nationalities, including coins and notes & 
        Indian Paper Currency Dataset (IPCD) with 50,263 images of 11 denominations including folded/partial notes & 
        Dongguk Korean Banknote v1 and Jordanian Dinar; 6,730 images of coins and banknotes in varied conditions & 
        USD, Euro, BDT; 30 classes, 25,012 augmented images, diverse backgrounds \\
        \hline
        \makecell{Results} & 
        Accuracy: 83.96\%, Precision: 83.96\%, Recall: 93.34\%, F1-Score: 88.40\% & 
        Accuracy: 96.75\% on IPCD; model optimized for low-resource devices (3.6M parameters) & 
        DKB v1: Precision: 97.21\%, Recall: 94.86\%, F1-Score;: 96.00\%; and for JOD: Precision: 96.04\%, Recall: 93.56\%, F1-Score: 94.78\% & 
        Accuracy: 97.73\%, Precision: 96.47\%, Recall: 95.23\%, F1-Score: 95.85\%, mAP@0.5: 97.21\% \\
        \hline
        \makecell{Processing Time} & 
        Not explicitly reported; designed for smartphone use & 
        Designed for resource-constrained smartphones; lightweight and efficient & 
        Desktop implementation using MATLAB with GPU; no real-time mobile deployment described & 
        Approx. 3.1 ms/image total inference time on smartphone \\
        \hline
        \makecell{Deployment} & 
        Model made publicly available; targeted smartphone implementation & 
        Android app "Roshni" with offline TensorFlow Lite model and voice feedback & 
        Desktop system; no mobile deployment detailed & 
        Mobile or web app with live voice feedback, Flask backend, and WebSocket frontend \\
        \hline
    \end{tabular}
    \caption{Comparison of the proposed model with recent studies on currency recognition.}
    \label{tab:recent_comparison}
    \end{threeparttable}
\end{table*}

\textbf{Methodology Comparison:}  
Park and Park (2023)\cite{math11061392} introduced a multinational banknote detection model (MBDM) employing a 69-layer convolutional neural network. They implemented mosaic data augmentation to improve generalization across currencies like Korean won, US dollar, Euro, and Jordanian dinar. The model is designed to handle small objects like coins with a specialized architecture but does not detail computational complexity. Singh et al. (2022)\cite{9868010} proposed IPCRNet, a lightweight neural network using a MobileNet front-end with a novel contextual block incorporating multi-dilation depthwise separable convolutions. This architecture reduces parameter count to approximately 3.6 million which makes it suitable for deployment on resource-constrained smartphones. Park et al. (2020)\cite{9217508} presented a three-stage detection framework based on Faster R-CNN and ResNet-18, using geometric constraints and multi-level verification to detect banknotes and coins. Their approach requires high computational resources and is implemented primarily on desktop GPUs. 

Our proposed YOLOv8n-based model leverages the nano variant with a custom detection head including Squeeze-and-Excitation blocks to automatically extract features, enabling adaptability across multiple currencies (USD, Euro, BDT) with improved speed and efficiency.

\textbf{Dataset Description:}  
The MBDM model by Park and Park (2023)\cite{math11061392} was trained and evaluated on a multinational dataset containing over 21,000 images spanning Korean won, US dollar, Euro, and Jordanian dinar currencies. The dataset includes coins and banknotes captured under diverse conditions. Singh et al. (2022)\cite{9868010} utilized the IPCD dataset, comprising more than 50,000 images of Indian paper currency across 11 denominations. They emphasized on real-world scenarios with folded and partial notes to simulate the perspective of visually impaired users. Park et al. (2020)\cite{9217508} tested their three-stage detection method on the Dongguk Korean Banknote database (6,400 images) and the Jordanian Dinar open database (330 images). They also focused on both coins and banknotes with varying background complexities. 

Our model is trained on a combined dataset of USD, Euro, and BDT currencies. We covered 30 classes with folded notes and extensive augmentation to increase robustness across diverse real-world scenarios.

\textbf{Results Comparison:}  
Park and Park (2023)\cite{math11061392} reported an overall accuracy of approximately 83.96\%, recall of 93.34\%, and an F1 score of 88.40\% on their multinational dataset, showing effective detection but lower than specialized single-country models. Singh et al. (2022)\cite{9868010} achieved an accuracy of 96.75\% on the IPCD dataset with a lightweight network architecture. They tried to balance accuracy and model size for mobile deployment. Park et al. (2020)\cite{9217508} demonstrated strong detection precision and recall on their banknote and coin datasets. Our proposed YOLOv8n-based model outperforms these approaches with an accuracy of 97.73\%, precision of 96.47\%, recall of 95.23\%, F1-score of 95.85\%, and a mean Average Precision at IoU=0.5 of 97.21\%, showing superior detection consistency and real-time applicability.

\textbf{Processing Time and Deployment:}  
Park and Park’s method (2023)\cite{math11061392} did not explicitly share the processing time. Singh et al. (2022)\cite{9868010} developed a lightweight CNN optimized for resource-constrained smartphones, deployed as an Android application named ``Roshni'' with offline TensorFlow Lite support, though exact per-image processing times are not provided. Park et al. (2020)\cite{9217508} implement a three-stage Faster R-CNN detection system primarily on a desktop with GPU acceleration, without details on real-time mobile deployment or processing speed. 

Our proposed YOLOv8n-based model processes each image in approximately 3.1 ms (0.2 ms preprocessing, 2.2 ms inference, 0.7 ms postprocessing) and is optimized for embedded deployment on smartphones and web applications, offering real-time detection with voice feedback suitable for practical scenarios such as ATMs and mobile devices.

\begin{figure*}[t]
    \centering
    \includegraphics[width=1\textwidth]{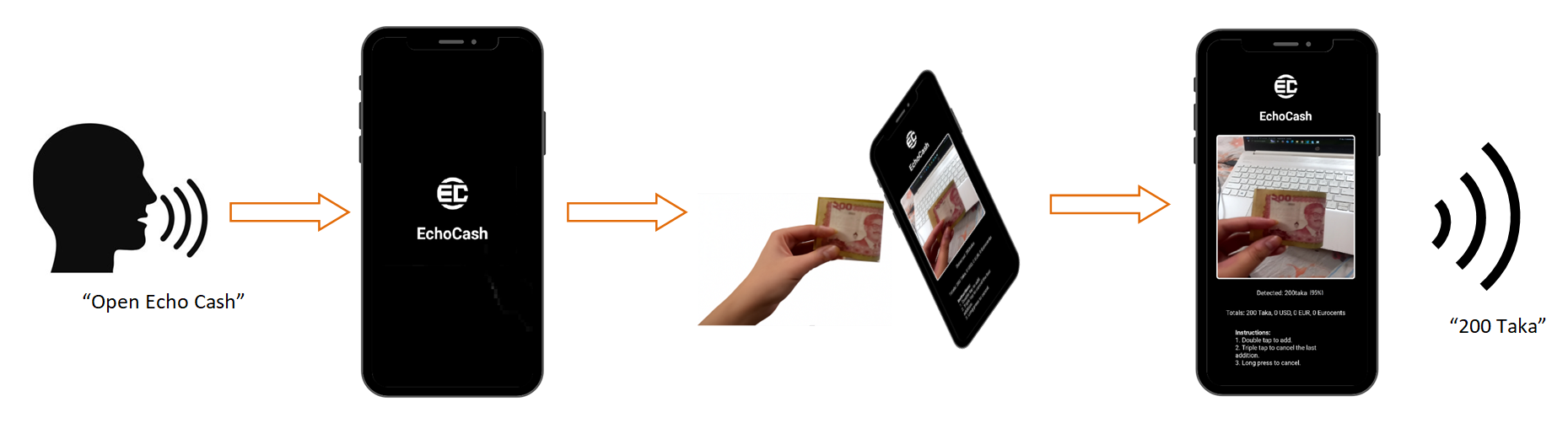}
    \caption{Proposed system and interface for the mobile app or web app, displaying detected currency with voice feedback option.}
    \label{fig:interface}
\end{figure*}

In summary, our YOLOv8n-based model surpasses Park and Park’s\cite{math11061392} multinational CNN approach in accuracy and real-time processing capability. It aligns with Singh et al.’s\cite{9868010} lightweight Indian currency recognition framework while improving inference speed and deployment flexibility and outperforms Park et al.’s\cite{9217508} three-stage Faster R-CNN detection system in practical applicability and efficiency. Future work could include integrating currency authenticity verification, addressing limitations highlighted in these prior works to further enhance the robustness and utility of assistive currency recognition systems.

\subsection{Deployment Strategy}

The proposed YOLOv8n-based model, optimized for real-time currency detection, is designed for deployment in a mobile application or web application to assist visually impaired users in identifying USD, Euro, and BDT currencies. Leveraging the model’s inference speed of 3.1 ms per image. The app is opened using voice assistant and the system utilizes smartphone camera to process currency images continuously, delivering rapid and reliable detection across 30 classes. A key feature of this deployment is the integration of voice feedback. The voice is activated when a currency denomination is consistently detected for 3 seconds. This temporal threshold ensures accuracy by requiring sustained recognition, mitigating errors from transient misclassifications or unstable inputs. Figure [\ref{fig:interface}] illustrates the proposed system and the interface of the app, showcasing a simple design where the detected currency is displayed with an option for voice output.

The 3-second detection window enhances reliability by allowing the model to confirm its prediction over multiple frames. It reduces the impact of brief occlusions, shaky hands or varying lighting conditions which are some common challenges in real-world mobile use. The confusion matrix (Figure [\ref{fig:confusion_matrix}]) shows minor misclassifications (e.g., "5euro" confused with "10euro" in 16 instances), which could occur momentarily due to visual similarities. By requiring consistent detection over 3 seconds, the system filters out such fleeting errors, ensuring the voice feedback—delivered via text-to-speech—accurately announces the correct denomination (e.g., "100 dollar" or "50 taka"). This delay balances speed and precision which is critical for visually impaired users relying on auditory cues for financial autonomy.

The system also features a tap-based interaction for counting the money. Once a currency denomination is detected and announced with voice feedback, a quick double-tap on the screen adds that amount to a running total and gives voice feedback. A triple-tap undoes the last addition with the system giving a voice feedback with the updated total. Pressing and holding for 1.5 seconds, it resets the total to zero and says "canceled." The totals are neatly tracked separately for USD, Euro (EUR), Eurocents, and Bangladeshi Taka (BDT). For every double-tap to add a denomination, the updated total is spoken aloud for confirmation. It keeps things clear and simple, especially with multiple currencies in play. Also the amount only gets added after the detection is locked in and the voice feedback has played, so accidental taps or detection won’t mess things up.

The system relies on cloud-based processing to deliver a seamless experience which makes it easy to update the model or currency dataset as needed. It is designed to empower visually impaired individuals with a straightforward, accurate and efficient way to identify currencies, all through an intuitive web interface.

\subsection{Discussion and Conclusion}
Our proposed YOLOv8n model with custom head for detection of USD, Euro, and BDT currencies delivers acceptable performance for 30 classes. 97.73\% accuracy, 96.47\% precision, 95.23\% recall, and 95.85\% F1-score were achieved with 50 epochs of exhaustive training. It shows the model's ability to identify currency denominations correctly with less false positives and missed detections. The confusion matrix (Figure [\ref{fig:confusion_matrix}]) also verifies the result by possessing large numbers of true positives along the main diagonal (e.g., 384 for "1000taka," 392 for "100dollar"), with very few misclassifications (e.g., "5euro" with 16 instances being mistaken for "10euro") owing to visual similarity in appearance or closeness of denomination. The training metrics and loss plots (Figure [\ref{fig:matrics_curve}]) show successful learning, with train and validation losses (box, class, and DFL) converging smoothly.  Performance metrics like precision and recall flattening at the level of 0.95, as expected for end results.

The model's mAP@0.5 measure of 97.21\% indicates excellent detection accuracy at a moderate localization threshold level. The mAP@0.5:0.95 measure of 80.38\% indicates robust performance across stricter IoU ranges suitable for real-world challenges like varying lighting and positions. One of its strengths is that its inference rate is 3.1 ms per image (0.2 ms preprocess, 2.2 ms inference, 0.7 ms postprocess). It outperforms the standard YOLOv8n’s 3.2 ms, making it highly practical for real-time deployment on resource-constrained devices such as smartphones, making it even more accessible for visually impaired users through rapid currency identification and voice feedback. The model could be employed within a smartphone application or web application for use of phone cameras or web interfaces to scan for currencies in real time, offering an easy utility for visually impaired consumers to manage transactions independently.

The use of deep convolutional layers with Squeeze-and-Excitation (SE) blocks in the custom detection head enhances its attention to currency-specific features. Additionally, the absence of authenticity detection (e.g., identifying counterfeit notes) is a practical limitation for broader applications of this system such as banking or commerce.

In conclusion, this model offers a balanced and efficient solution for multi-currency recognition with an accuracy of 97.73\%, precision of 96.47\%, recall of 95.23\% and real-time processing capability at 3.1 ms per image which meets the need of assistive technologies. Its integration in a mobile app or web app can be accessible for visually impaired users to use a real-time currency detection system. Its training stability and high detection accuracy underscore the efficacy of the YOLOv8n framework for this task. Future scopes include expanding the dataset with quantified samples to enhance transparency and generalization, integrating counterfeit detection to broaden utility and testing on additional currencies to improve scalability and support a wider range of global currencies. Fine-tuning the model could further narrow the training-validation loss gap which may potentially boost mAP@0.5:0.95 beyond 80.38\%. Exploring lightweight architectures could reduce inference time below 3.1 ms which will enhance the integration on low-end devices. Such enhancements may make the model a reliable tool for wide-area global currency recognition in a range of real-world scenarios.



\bibliographystyle{ieeetr}

\end{document}